\documentclass[conference]{IEEEtran}
\IEEEoverridecommandlockouts
\usepackage{cite}
\usepackage{amsmath,amssymb,amsfonts}
\usepackage{algorithmic}
\usepackage{graphicx}
\usepackage{textcomp}
\usepackage{xcolor}

\usepackage{booktabs}
\usepackage{comment}
\usepackage{adjustbox}
\usepackage{multirow}
\usepackage[ruled,vlined]{algorithm2e}
\def\BibTeX{{\rm B\kern-.05em{\sc i\kern-.025em b}\kern-.08em
    T\kern-.1667em\lower.7ex\hbox{E}\kern-.125emX}}
\begin{document}

\title{The Effects of Grouped Structural Global Pruning of Vision Transformers on Domain Generalisation\\
}
\author{Hamza Riaz and~Alan~F. Smeaton
\thanks{H Riaz and A.F. Smeaton are with the School of Computing, Dublin City University, Ireland.}
}


\maketitle

\begin{abstract}
With the growing sizes of AI models like large language models (LLMs) and vision transformers, deploying them on devices with limited computational resources is a significant challenge particularly when addressing domain generalisation (DG) tasks. This paper introduces a novel grouped structural pruning method for pre-trained vision transformers (ViT, BeiT, and DeiT), evaluated on the PACS and Office-Home DG benchmarks. Our method uses dependency graph analysis to identify and remove redundant groups of neurons, weights, filters, or attention heads within transformers, using a range of selection metrics. Grouped structural pruning is applied at pruning ratios of 50\%, 75\% and 95\% and the models are then fine-tuned on selected distributions from DG benchmarks to evaluate their overall performance in DG tasks. 
Results show significant improvements in inference speed and fine-tuning time with minimal trade-offs in accuracy and DG task performance. For instance, on the PACS benchmark, pruning ViT, BeiT, and DeiT models by 50\% using the Hessian metric resulted in accuracy drops of only -2.94\%, -1.42\%, and -1.72\%, respectively, while achieving speed boosts of 2.5x, 1.81x, and 2.15x. These findings demonstrate the effectiveness of our approach in balancing model efficiency with domain generalisation performance. 
\end{abstract}


\section{Introduction}

As a consequence of  recent  developments in the field of AI especially in generative models and large language models (LLMs), the compression of deep learning models has became an important topic. To deploy large models on edge devices, compression methods for neural networks have been developed as summarised in~\cite{yu2023dataset} which is sometimes  referred to as optimisation or quantisation. Pruning has also become highly effective and practical compared to other compression paradigms~\cite{liang2021pruning}. The goal of pruning a neural network is to remove  redundant parameters which reduces its size thus improving the inference speed of models. For example, the widely used ResNet-50~\cite{he2016deep} requires about 95 MB of RAM and has over 23 million parameters~\cite{youdrawing}. For models like $BERT_{BASE}$~\cite{devlin-etal-2019-bert}, the size is around 440 MB with 110 million parameters, GPT-3 contains up to 175 billion parameters~\cite{wu2023ai}, GPT-4 has even more. The trend of developing and releasing larger models has become a race among big technology companies and research groups creating the need for pruning methods which could be used during or after development of  models. 

It is clear that deep neural networks (DNNs) often require significant time and memory for processing, posing challenges for deployment on devices with limited computational resources such as those in real-time applications like autonomous driving. Furthermore, DNNs with redundant features and high-dimensional feature spaces might present more entry points for adversarial attacks, compromising the network's capacity to generalise beyond its initial training data~\cite{sehwag2020hydra}. All of this has led to increased interest in neural network compression techniques, such as pruning~\cite{ashkboos2024slicegpt}, low rank factorisation~\cite{denton2014exploiting}, quantisation~\cite{dettmers2024qlora}, neural architecture search~\cite{liu2018darts}, and knowledge distillation~\cite{xu2024survey} which  create lightweight models that reduce memory and computational demands while maintaining or improving performance. 

Although pruning approaches can be divided into different schemes,   we consider only mainstream pruning approaches namely structural pruning~\cite{you2019gate} and unstructural pruning~\cite{dong2017learning}. 

The primary distinction between the two is that structural pruning alters the structure of neural networks by  deleting grouped parameters, whereas unstructural pruning zeros partial weights without modifying the network structure. In reality, for small or medium models, unstructured pruning often sets their respective masks (or indicators) $m$ to 0, rather than the weights themselves. Assigning such a binary mask to each weight in large models such as LLMs is difficult because of the enormous number of weights~\cite{wang2020picking}. One of the  motivations for using structural pruning  is that it does not need any  extra support from external hardware or software to deploy  as it can reduce size and speed up networks directly~\cite{li2016pruning}. 

Domain generalisation (DG) addresses a scenario where a model $M_1$ trained on a source domain $D_1$ is evaluated on an unseen target domain $D_2$. The performance drop observed due to the differences between $D_1$ and $D_2$ is attributed to \textit{domain shift}, a phenomenon also known as the \textit{out-of-distribution (OOD) problem}. DG aims to address this challenge by enabling a model to pool knowledge from one or more source domains into a unified representation that can generalise effectively to unseen domains without requiring additional retraining.
DG gives the ability to models to generalise and to perform well on unseen data domains or distributions, which is a common challenge in machine learning when handling domain shifting \cite{wang2022generalizing}.

There is a significant gap in the existing literature which is the lack of studies on the relationship between network pruning and DG. 
In this paper, we implemented structural pruning methods on three  vision transformer models --  ViT, BeiT and DeiT -- for two DG benchmarks. Inspired from~\cite{fang2023depgraph}, the pruning mechanisms consist of finding a dependency graph and then grouping and group-level pruning for the selected model. The  structural pruning removes sets of parameters scattered over many layers. The parameters within each group are inter-dependent because of layer-to-layer connections, therefore they must be pruned simultaneously to maintain the model's structural integrity. The method employs dependency graphs to automatically recognise these relationships and to gather parameter groupings for pruning~\cite{fang2023depgraph}. It then prunes  pre-trained models from base settings into new models which will be structurally different from the originals, and we can then fine-tune the  pruned models for DG benchmarks.

The structure of this paper has an introduction and related work sections followed by the motivation for using structural pruning. A  methodology section shows how we implemented structural pruning for  vision transformers to explore the effects  on DG benchmarks. The results section discusses the trade-offs in speed, time and accuracy, with a concluding section summarising the paper.

\section{Related Work}

In this paper we divide  previous work on pruning network models into three types, namely pruning after,  during, and before training.  

\subsection{Pruning Before training}
Pruning before training sparsifies the structure of a neural network before starting the training process. Instead of training a whole network and then pruning after or during training, the model is trimmed at the start or during early phases of training. This minimises the model's size from the start which may result in more efficient training and inference while preserving model performance.




One of the important works in early-stage pruning of neural networks is known as the ``Lottery Ticket Hypothesis"~\cite{frankle2018lottery}. This proposes that within a randomly initialised dense neural network, there exists a smaller, sparse subnetwork (a ``winning ticket") that can be trained to match the performance of the original dense network/models when trained in isolation. This approach includes training the network once, pruning, resetting the pruned weights to their initial values, and then retraining the sparse network to determine the winning ticket. This technique  requires iterative pruning and retraining which can be computationally costly. As a refinement of the lottery ticket hypothesis, single-shot network pruning (SNIP) has been proposed~\cite{lee2018snip} where we have some randomly initialised mesh and we leave the most promising connections in it. These are determined by their connection sensitivity referring to how much the removal of a specific weight affects the loss so that the least influential ones are removed. While computationally efficient, SNIP struggles to prune large percentages of weights~\cite{lee2018snip}. To extend the functionality of SNIP, the gradient signal preservation (GraSP) method was developed~\cite{wang2020picking}. When we trim a model before training, it generate a sparse structure of neural networks which are challenging to train.  GraSP enhances SNIP by focusing not just on weight significance but also on preserving the gradient flow after pruning, which assists in training stability. 
Because of its emphasis on gradient preservation, GraSP requires more processing  than SNIP~\cite{wang2020picking}. 

Existing gradient-based pruning strategies which eliminate portions of a neural network at the start of training can occasionally result in layer collapse. Layer collapse occurs when an entire layer of the network is mistakenly pruned (removed) rendering the network incapable of learning or functioning properly. A  pruning method called iterative synaptic flow pruning (SynFlow) avoids layer-collapse~\cite{tanaka2020pruning} by examining the synaptic flow of information throughout network, while pruning. 

\subsection{Pruning During Training}
Pruning during training refers to progressively removing the least important weights or neurons from a neural network while it is being trained. Such methods allow the models to dynamically adjust and recover from pruning during the training process.

Mathematically, we consider the weight matrix at a given training step \( t \) as \( W_t \). A binary mask \( M_t \in \{0, 1\}^{d_1 \times d_2} \) is applied to the weight matrix, where the mask determines which weights remain active. The effective weight matrix at step \( t \), after pruning, is:

\[
\tilde{W}_t = W_t \odot M_t
\]

\noindent 
where \( \odot \)  denotes element-wise multiplication. During training, the mask \( M_t \) is updated based on a pruning criterion such as weight magnitude or gradients. The training process aims to minimise the loss function:

\[
L(W_t \odot M_t)
\]

\noindent 
subject to a sparsity constraint:

\[
\sum_{i,j} M_{i,j} \leq s
\]

\noindent 
where \( s \) denotes the desired level of sparsity, i.e. the number of active weights to maintain during training. 
Typically this begins with a randomly initialised dense network as the input model and simultaneously trains and prunes the network by updating both the weights and the associated masks which may represent weights, filters, channels, etc. The approaches described in for example~\cite{evci2020rigging, he2018soft, cho2024pdp} are well-known for pruning during training. Methods for pruning during training can be divided into four  types namely  sparsity regularisation based~\cite{gordon2018morphnet}, dynamic spare training based~\cite{evci2020rigging}, score-based~\cite{he2018soft, wang2020picking}, and differentiable pruning based methods~\cite{ning2020dsa, cho2024pdp}.

\subsection{Pruning After Training}
Pruning after training is the practice of eliminating redundant or less significant parameters from a neural network once it has been fully trained. Most pruning methods~\cite{molchanov2016pruning, li2016pruning,  dong2017learning, NIPS1989_6c9882bb} work with a pre-trained network. The fundamental concept here is to identify which weights are most redundant and, as a result, will have least impact on performance when removed. Magnitude-based pruning  eliminates weights that are less than a threshold, which may miscalculate the relevance of each weight. In contrast, Hessian-based pruning  determines the significance of each weight by calculating how its removal would effect the loss~\cite{NIPS1989_6c9882bb}. However, all  the aforementioned solutions need pre-training and hence are not relevant at initialisation.

Pruning after training is a popular strategy in neural network optimisation that  can also be divided into four types including magnitude-based, iterative, structured, and dynamic pruning, each with its advantages and disadvantages. Magnitude-based approaches are straightforward but have an uneven effect on layers whereas iterative pruning allows for performance recovery. Structured pruning maximises hardware efficiency, whereas dynamic pruning provides flexibility during inference. According to the literature, the proper pruning strategy is ultimately determined by the unique use case, model architecture, and deployment environment~\cite{10643325}.

\section{Methodology} \label{method}
Our method to prune  vision transformers is inspired by~\cite{fang2023depgraph} and follows the following steps.

\subsection{Finding Dependency for Networks}

The first stage in the process is to analyse the  network to discover which components (nodes) are interdependent. 
The  network is represented as a Dependency Graph~\cite{fang2023depgraph} with each layer, filter, or channel  viewed as a node  and the relationships between them are edges.  
This dependence analysis is crucial since trimming one element of the network may influence another. 
We consider a neural network \( N \) which can be represented as a set of layers, filters, or channels:
\[
N = \{L_1, L_2, ..., L_n\}.
\]
Each element of the network has dependencies on other elements in terms of connections, which can be fully connected layers, skip connections, concatenation or residual connections. This can be presented as a dependency graph \( G = (V, E) \), where:
\begin{itemize}
    \item \( V = \{v_1, v_2, ..., v_n\} \) refers as a set of nodes of network components (like filters, channels, etc.),
    \item \( E \) is the set of directed edges representing dependencies between nodes.
\end{itemize}

Once the dependency graph has been constructed, a pruning approach is implemented where groups of dependent parameters are trimmed simultaneously.  In {Dynamic Pruning} the designed framework  motivated by~\cite{fang2023depgraph} supports  pruning across many structures (filters, channels, and layers) and modifies the network structure accordingly. Pruning is not confined to a particular type of structure, making it very adaptable. In our case we  prune the structure of pre-trained models and then perform fine-tuning for domain generalisation. 

Each node \( v_i \) has a score \( S(v_i) \) which quantifies its importance (e.g., weight magnitude, or gradient). In our  approach, we use 5 types of importance score metrics including, $L_1$ norm, $L_2$ norm, Taylor, random, and Hessian. These metrics set a threshold \( \theta \), and nodes with scores below this threshold are pruned.
Due to dependencies, when a node \( v_i \) is pruned all nodes \( v_j \) dependent on \( v_i \) must also be pruned as defined by the edges in the graph \( (v_i, v_j) \in E \).
The pruning process is formalised as: for each node $v_i \in V$, if  $S(v_i) < \theta$,  prune  $v_i$  and all nodes  $v_j$  such that $(v_i, v_j) \in E$.

The above  ensures that if a component/element is selected for pruning then all dependent components/elements also are pruned, preserving the structural consistency of the network.
For dynamic pruning this applies to different types of structure in the network. For example, a filter \( F \), a channel \( C \), or an entire layer \( L \) may be treated as nodes in the dependency graph. The pruning criterion applies to each structure \( s \):
\[
\text{Prune structure } s \text{ if } S(s) < \theta,
\]
where \( s \in \{F, C, L\} \).

Our approach then further refines the dependency graph once it has been created and pruned after performing the above process. The objective is to reduce network size and maintain  performance. Fine-tuning involves retraining the model after pruning to recover accuracy lost throughout the process.

Let the pruned network be represented by \( N_p \). The training objective is to minimise the loss \( Loss \) over \( N_p \):
\[
\min_{W_p} Loss(N_p(W_p)) \quad \text{subject to} \quad \|W_p\|_0 \leq \kappa,
\]
where \( W_p \) expresses the weights of \( N_p \) and \( \kappa \) is a constraint on the number of remaining parameters.

\subsection{Pruned Vision Transformer Models Fine-Tuning on Domain Generalisation Benchmarks}

In our experiments  we use to popular DG benchmarks  PACS \cite{li2017deeper} and Office-Home \cite{venkateswara2017Deep}.
For our investigation into pruning of vision transformers, we perform experiments with three transformer models namely ViT, BeiT and DeiT with  variations as described later. 

Our framework runs an importance score criterion on each model with selected hyperparameters and benchmarks. The pruned weights and structures are then stored for each  and the pruning weights are further fined-tuned for evaluation on domain generalisation benchmarks. In addition, it is important to note that the pruning step and the fine-tuning of the models will have different sets of hyperparameters.

\subsection{Inference of New Fine-Tuned and Pruned Models}

After performing pruning of pre-trained weights and fine-tuning  these  according to our DG benchmarks, the final step is to evaluate the performance of the pruned models on test distributions. We divided each DG benchmark into a 80\% / 20\% ratio of training+validation and testing. For instance, in the case of Office-Home, we used Art, Product and Real World domains in training+validation and Clipart in testing. 

\begin{figure}[!htb]
\centering
\includegraphics[width=\columnwidth]{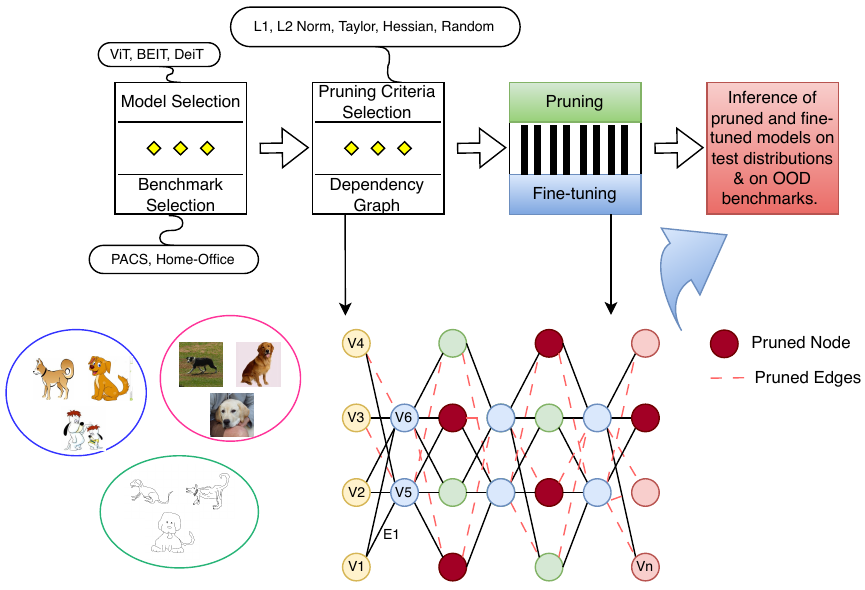} %
\caption{Overview of group structural pruning for vision transformers.
\label{chapter7_fig1}}
\end{figure}

Figure~\ref{chapter7_fig1} presents an overview of our experiments on  structural pruning.

\section{Experiments}
We converted images from the domain generalisation benchmarks PACS \cite{li2017deeper} and Office-Home \cite{venkateswara2017Deep} into ImageNet data format. 
For working with the three vision transformer models  we  used base versions  vit-base-patch16-224, beit-base-patch16-224, and deit-base-distilled-patch16-224 respectively. 
After initialisation of weights, we applied group pruning based on importance score metrics namely L1 and L2 Norm~\cite{li2016pruning}, Taylor~\cite{molchanov2019importance}, Hessian~\cite{NIPS1989_6c9882bb}, and random. In  group structural pruning we removed groups of filters, attention heads, channels, or layers of models and the elimination rate was decided by the importance metrics. For instance, L1 and L2 Norm are motivated from~\cite{li2016pruning} where the authors explain that usually both norms have similar response in the selection process with only small differences. Similarly, the Taylor importance estimation is inspired by~\cite{molchanov2019importance} which implemented first-order and second-order of Taylor importance on filters or neurons. Results in~\cite{molchanov2019importance} showed that both methods are extremely close in terms of performance therefore we only implemented the first-order Taylor importance method.  Hessian group pruning is also known as optimal brain damage (OBD)~\cite{NIPS1989_6c9882bb} and uses information-theoretic principles to determine which weights may be deleted with lowest impact on performance.  Finally, in the random group selection we pick and drop groups of (neurons) randomly without any importance selection mechanism to check the general performance of sub-networks already present in the pre-trained weights.

As  mentioned in Section~\ref{method}, there are two major steps  in the development namely pruning of vision transformers and fine-tuning the pruned models. The pruning process has hyperparameters including pruning ratio, number of heads, dimensions of output heads, and global pruning. The term ``global pruning'' means it takes the entire network into account and prunes weights across all layers by applying a  balanced reduction of network size whereas local pruning focuses on the elimination of individual layers or specific sections of the network using criteria such as weight magnitude. 

For our experiments  we implemented global pruning with three variations in pruning ratio, 50\%, 75\%, and 95\%. The number of heads are reduced from 144 to 72 with each head's dimensionality as 64. This settings is applied on all  models with selected importance score metrics and for each base model we have 15 pruned variations and the performance results for this are shown later.

In the fine-tuning of stored pruned architectures of models, we have a range of  hyperparameters such as epoch, batch size, optimiser, initial learning rate, momentum, weight-decay, learning rate scheduler, learning rate warm up method, and learning rate warm up-decay. The fine-tuning process will continue in each experiment for 300 epochs, with AdamW as the optimiser, 0.00015 as the initial learning rate, 0.9 as momentum, 0.3 as weight-decay, CosineAnnealingLR as the learning rate scheduler, linear as learning rate warm up method, and 0.033 as the learning rate warm up decay. We did not implement sparse or dynamic training and pruning methods for vision transformers.

\section{Results and Analysis}

\begin{table*}[!htb]
\caption{Comparison of base model performances on key metrics, namely the number of trainable parameters, computational cost (MACs), accuracy, and fine-tuning time for PACS \& Office-Home.
\label{chapter7_tab1}}
\centering
\begin{adjustbox}{width=0.85\textwidth} 
\renewcommand{\arraystretch}{0.75}
\begin{tabular}{lcc|ccc|ccc}
\toprule
\textbf{Model Name} & \textbf{\# Params.} & \textbf{MACs} & \multicolumn{3}{c|}{\textbf{PACS}} & \multicolumn{3}{c}{\textbf{Office-Home}} \\
\cmidrule(lr){4-6} \cmidrule(lr){7-9}
 & & & \textbf{Valid Acc. (\%)} & \textbf{Test Acc. (\%)} & \textbf{FT-Time} & \textbf{Valid Acc. (\%)} & \textbf{Test Acc. (\%)} & \textbf{FT-Time} \\
\midrule
ViT-Base   & 86.57~M & 17.59G & 94.43 & 94.33 & 04:00:54 & 86.48 & 51.088 & 06:25:18 \\
BEIT-Base  & 86.53~M & 12.67G & 96.41 & 95.64 & 04:10:24 & 87.75 & 55.052 & 06:39:55 \\
Deit-Base  & 87.34~M & 17.69G & 96.87 & 96.56 & 04:01:35 & 86.82 & 51.340 & 06:25:11 \\
\bottomrule
\end{tabular}
\end{adjustbox}
\end{table*}

Table~\ref{chapter7_tab1} shows performance for the base versions of three vision transformers and has nine columns  headed  model name, \# Params which is the total number of trainable parameters in each model  expressed in millions, MACs shows the number of operations required to perform a single forward pass measured in Giga (G) operations (lower is better), validation and test accuracies, and fine-tuning time measured in hours, minutes, and seconds (HH:MM:SS) for each of the PACs and Office-Home benchmarks.  
Higher validation accuracy means higher generalisation to unseen data after training while test accuracy measures the performance of models on the test dataset/distribution and  is another indicator of how effectively a model generalises. 
Table \ref{chapter7_tab1} shows that Deit-Base performs  best in terms of validation and test accuracies, while having the highest number of parameters and MACs reflecting a vital insight already established in \cite{riaz2023domain}, that bigger models usually have better domain generalisation capabilities. On the other hand, BEIT-Base is the most computationally efficient (lowest MACs)  yet providing excellent performance. ViT-Base has less accuracy than the others although it is comparable to Deit-Base in terms of size and processing needs.

\subsection{Group Structural Pruning Results for  ViT Transformer}

\begin{table*}[!htb]
\caption{Pruning methods on the ViT model at various pruning ratios.\label{chapter7_tab2}}
\centering
\begin{adjustbox}{width=0.85\textwidth} 
\renewcommand{\arraystretch}{0.75}
\setlength{\tabcolsep}{2pt} 
\begin{tabular}{p{1.5cm}p{1.8cm}p{1.5cm}p{1.5cm}p{1.8cm}p{1.8cm}p{1.5cm}p{1.2cm}p{1.2cm}}
\toprule
\textbf{Pruning ~~ Ratio} & \textbf{Group Score} & \textbf{\#~Pruned Params} & \textbf{\#~Pruned MACs} & \textbf{Valid Acc (\%)} & \textbf{Test Acc (\%)} & \textbf{$\Delta$Acc (\%)} & \textbf{Speed Up} & \textbf{FT-Time} \\
\midrule
\multirow{5}{*}{ViT-50\%} 
& Hessian & 35.03M & 6.94G & 93.50 & 91.39 & -2.94 & 2.5x & 01:59:08 \\
& Taylor & 33.68M & 6.66G & 87.01 & 84.70 & -9.63 & 2.6x & 02:00:45 \\
& Random & 21.01M & 4.45G & 78.49 & 78.22 & -16.11 & 3.95x & 01:32:15 \\
& L1-Norm & 34.85M & 6.88G & 77.62 & 73.76 & -20.57 & 2.55x & 02:04:31 \\
& L2-Norm & 31.03M & 6.21G & 77.22 & 75.38 & -18.95 & 2.83x & 01:58:12 \\
\midrule
\multirow{5}{*}{ViT-75\%} 
& Hessian & 12M & 2.72G & 85.10 & 83.79 & -10.54 & 6.5x & 01:15:18 \\
& Taylor & 6.19M & 1.75G & 70.55 & 70.11 & -24.22 & 10.05x & 00:55:23 \\
& Random & 7.02M & 1.91G & 69.62 & 66.87 & -27.46 & 9.21x & 01:02:32 \\
& L1-Norm & 30.45M & 5.84G & 77.56 & 75.18 & -19.15 & 3.01x & 01:49:16 \\
& L2-Norm & 4.98M & 1.56G & 67.25 & 66.87 & -27.46 & 11.28x & 00:55:22 \\
\midrule
\multirow{5}{*}{ViT-95\%} 
& Hessian & 0.60M & 0.83G & 52.93 & 50.15 & -44.17 & 21.19x & 00:42:12 \\
& Taylor & 0.11M & 0.73G & 36.00 & 32.22 & -62.11 & 24.10x & 00:35:21 \\
& Random & 1.03M & 0.89G & 62.44 & 62.01 & -32.32 & 19.76x & 00:46:00 \\
& L1-Norm & 18.68M & 3.52G & 76.23 & 73.76 & -20.57 & 5.0x & 01:24:32 \\
& L2-Norm & 0.21M & 0.76G & 46.73 & 43.97 & -50.35 & 23.14x & 00:41:28 \\
\bottomrule
\end{tabular}
\end{adjustbox}
\end{table*}
We now present detailed analysis of our experiments to explore  domain generalisation with pruning on vision transformers. Table \ref{chapter7_tab2} shows pruning results for the ViT model at three  pruning ratios, 50\%, 75\%, and 95\%. The table compares the impact of several pruning strategies (Hessian, Taylor, Random, L1-Norm, and L2-Norm) and has 9 columns namely pruning ratio, group importance score, number of pruned parameters, number of pruned MACs, pruned validation and testing accuracies, $\delta$Acc which is the measure of change in accuracy and  calculated by subtracting pruned test accuracy from base test accuracy from Table~\ref{chapter7_tab1}, speedup and fine-tuning time. 

At 50\% pruning, the \textbf{Hessian-based} method achieves the best performance for the ViT model, delivers validation accuracy of 93.5\%, test accuracy of 91.39\% (with only a -2.94\% decline), and 2.5x speedup. This corresponds to a significant reduction in model size, from 86.57 M to 35.03 M parameters, and MACs from 17.59G to 6.64G. In contrast, the \textbf{Taylor-based} method shows a more substantial performance drop, with a test accuracy of 84.70\% (-9.63\%) but compensates with a slightly higher speedup of 2.6x. Random pruning provides the highest speedup at 3.95x but at the expense of steep performance decline (-16.11\%). \textbf{L1-Norm -based} and \textbf{L2-Norm-based} methods similarly result in lower accuracy (73.76\% and 78.22\%, respectively), with speedups ranging from 2.5x to 3.95x. These results demonstrate that while \textbf{Hessian-based} is better at preserving performance at moderate pruning levels, other methods  focus on computational gains at the cost of accuracy.

As pruning rates increase, trade-offs become more pronounced. At 75\% pruning, \textbf{Hessian} retains strong performance, with an accuracy of 83.79\% (-10.54\%) and 6.5x speedup, outperforming other methods in accuracy while maintaining efficiency. 
At extreme pruning rates (95\%), accuracy losses are severe across all methods, with Hessian and L1-Norm retaining better performance compared to \textbf{Taylor} and \textbf{Random}. L1-Norm stands out for its stable accuracy (73.76\%) and comparatively smaller losses, attributed to its ability to maintain larger structures post-pruning. Interestingly, \textbf{Random} pruning at high rates suggests potential for discovering sub-networks that retain reasonable accuracy (62.01\%) with substantial speedup (19.76x). Overall, Hessian consistently excels at moderate pruning levels, while L1-Norm provides a balance of accuracy and efficiency, particularly at higher pruning ratios.

\subsection{Group Structural Pruning Results for BEIT Transformer}

\begin{table*}[!htb]
\caption{Pruning methods on the BEIT model at various pruning ratios.\label{chapter7_tab3}}
\centering
\begin{adjustbox}{width=0.85\textwidth} 
\renewcommand{\arraystretch}{0.75}
\setlength{\tabcolsep}{2pt} 
\begin{tabular}{p{1.5cm}p{1.8cm}p{1.5cm}p{1.5cm}p{1.8cm}p{1.8cm}p{1.5cm}p{1.2cm}p{1.2cm}}
\toprule
\textbf{Pruning ~~ Ratio} & \textbf{Group Score} & \textbf{\# Pruned Params} & \textbf{\# Pruned MACs} & \textbf{Valid Acc (\%)} & \textbf{Test Acc (\%)} & \textbf{$\Delta$Acc (\%)} & \textbf{Speed Up} & \textbf{FT-Time} \\
\midrule
\multirow{5}{*}{BEIT-50\%} 
& Hessian & 57.79M & 7.01G & 94.96 & 94.23 & -1.42 & 1.81x & 03:11:36 \\
& Taylor & 58.03M & 7.06G & 94.15 & 93.52 & -2.13 & 1.80x & 03:12:06 \\
& Random & 58.17M & 7.09G & 93.74 & 92.91 & -2.74 & 1.79x & 03:11:37 \\
& L1-Norm & 57.88M & 7.03G & 92.52 & 90.58 & -5.07 & 1.80x & 03:11:53 \\
& L2-Norm & 58.43M & 7.14G & 92.70 & 89.97 & -5.67 & 1.77x & 03:12:38 \\
\midrule
\multirow{5}{*}{BEIT-75\%} 
& Hessian & 44.13M & 4.32G & 92.99 & 92.10 & -3.55 & 2.93x & 02:42:14 \\
& Taylor & 44.01M & 4.30G & 91.30 & 90.78 & -4.86 & 2.95x & 02:41:42 \\
& Random & 44.05M & 4.30G & 89.91 & 89.36 & -6.28 & 2.95x & 02:42:04 \\
& L1-Norm & 44.29M & 4.35G & 88.29 & 87.44 & -8.21 & 2.91x & 02:42:19 \\
& L2-Norm & 44.07M & 4.31G & 88.58 & 87.94 & -7.70 & 2.94x & 02:42:09 \\
\midrule
\multirow{5}{*}{BEIT-95\%} 
& Hessian & 32.71M & 2.07G & 83.19 & 82.78 & -12.87 & 6.12x & 02:16:57 \\
& Taylor & 32.74M & 2.08G & 86.90 & 85.11 & -10.54 & 6.09x & 02:17:00 \\
& Random & 32.68M & 2.06G & 79.25 & 78.32 & -17.33 & 6.15x & 02:17:01 \\
& L1-Norm & 32.75M & 2.08G & 83.54 & 83.08 & -12.56 & 6.09x & 02:16:47 \\
& L2-Norm & 32.61M & 2.05G & 83.19 & 80.14 & -15.50 & 6.18x & 02:16:34 \\
\bottomrule
\end{tabular}
\end{adjustbox}
\end{table*}

Table \ref{chapter7_tab3}  has the same  evaluation metrics as those in Table~\ref{chapter7_tab2}. At a 50\% pruning ratio, the \textbf{Hessian} technique demonstrates the best performance on the PACS benchmark, maintaining a pruned test accuracy of 94.23\% with a minimal accuracy loss of -1.42\%. This is achieved by reducing 57.79 M parameters and cutting MACs to 7.01G, with a moderate 1.81x speedup and reduced fine-tuning time of 03:11:36. Hessian prioritises accuracy retention over speed, making it ideal for applications requiring high precision. Taylor pruning follows closely with a test accuracy of 93.52\% and a similar speedup of 1.80x, albeit slightly less accurate than Hessian. Random pruning is less effective, achieving an accuracy of 92.91\% with a larger loss of -2.74\%, while L1-Norm and L2-Norm perform poorest, with accuracies of 90.58\% and 89.97\%, respectively, and losses exceeding -5\%. The reliance of L1-Norm and L2-Norm on magnitude-based weight selection appears inadequate for capturing critical weights in a complex model like BEIT.

At 75\% pruning ratio, \textbf{Hessian-base} remains the most effective, achieving a pruned test accuracy of 92.10\% with an accuracy reduction of -3.55\% and a significant speedup of 2.93x. Other methods follow similar trends to those seen at 50\%, with accuracies ranging between 87\%-91\%, still close to the baseline BeiT model. At the extreme 95\% pruning ratio, Taylor outperforms other methods with the highest pruned accuracy of 85.11\% and a speedup of 6.09x, indicating its gradient-based scoring becomes more effective as the network shrinks. 
Overall, \textbf{Hessian-base} excels across all pruning ratios due to its curvature-based scoring, while Taylor gains competitiveness at extreme pruning levels, and \textbf{Random-based}, \textbf{L1-Norm-based}, and \textbf{L2-Norm-based} lag in preserving accuracy under heavy pruning.

\subsection{Group Structural Pruning Results for DeiT Transformer}

\begin{table*}[!htb]
\caption{Pruning methods on the DeiT model at various pruning ratios.\label{chapter7_tab4}}
\centering
\begin{adjustbox}{width=0.85\textwidth} 
\renewcommand{\arraystretch}{0.75} 
\setlength{\tabcolsep}{2pt} 
\begin{tabular}{p{1.5cm}p{1.8cm}p{1.5cm}p{1.5cm}p{1.8cm}p{1.8cm}p{1.5cm}p{1.2cm}p{1.2cm}}
\toprule
\textbf{Pruning ~~ Ratio} & \textbf{Group Score} & \textbf{\# Pruned Params} & \textbf{\# Pruned MACs} & \textbf{Valid Acc (\%)} & \textbf{Test Acc (\%)} & \textbf{$\Delta$Acc (\%)} & \textbf{Speed Up} & \textbf{FT-Time} \\
\midrule
\multirow{5}{*}{DEIT-50\%} 
& Hessian & 43.02M & 8.24G & 95.30 & 94.83 & -1.72 & 2.15x & 02:12:50 \\
& Taylor & 38.05M & 7.36G & 92.64 & 90.27 & -6.28 & 2.40x & 02:12:06 \\
& Random & 22.20M & 4.62G & 83.30 & 82.98 & -13.58 & 3.83x & 01:28:22 \\
& L1-Norm & 32.13M & 6.34G & 81.97 & 80.85 & -15.70 & 2.79x & 01:58:15 \\
& L2-Norm & 34.62M & 6.77G & 87.77 & 86.12 & -10.44 & 2.61x & 01:59:14 \\
\midrule
\multirow{5}{*}{DEIT-75\%} 
& Hessian & 8.01M & 2.04G & 77.86 & 75.89 & -20.67 & 8.67x & 00:57:21 \\
& Taylor & 4.94M & 1.54G & 72.81 & 72.14 & -24.42 & 11.49x & 00:50:31 \\
& Random & 8.47M & 2.16G & 71.88 & 70.21 & -26.34 & 8.19x & 01:00:57 \\
& L1-Norm & 0.35M & 0.79G & 40.52 & 37.29 & -59.27 & 22.39x & 00:39:05 \\
& L2-Norm & 0.67M & 0.84G & 51.19 & 47.92 & -48.63 & 21.06x & 00:39:10 \\
\midrule
\multirow{5}{*}{DEIT-95\%} 
& Hessian & 0.47M & 0.81G & 58.15 & 54.41 & -42.15 & 21.84x & 00:38:13 \\
& Taylor & 0.12M & 0.75G & 34.90 & 32.02 & -64.54 & 23.59x & 00:35:37 \\
& Random & 1.03M & 0.88G & 64.87 & 64.13 & -32.42 & 20.10x & 00:37:49 \\
& L1-Norm & 0.07M & 0.74G & 17.10 & 17.33 & -79.23 & 23.91x & 00:36:59 \\
& L2-Norm & 0.07M & 0.74G & 17.10 & 17.33 & -79.23 & 23.91x & 00:36:53 \\
\bottomrule
\end{tabular}
\end{adjustbox}
\end{table*}

Table \ref{chapter7_tab4} indicates that at 50\% pruning, the Hessian approach surpasses the other strategies by keeping its test accuracy at 94.83\% while incurring just -1.72\% decrease in accuracy. With 43.02M trimmed parameters and 8.24G MACs, it achieves a 2.15x speedup for this accuracy. When we compare it with Hessian in BeiT at a similar pruning ratio it actually performs better in terms to speed and accuracy while the Taylor approach is similarly successful but suffers -6.28\% accuracy loss which reduces test accuracy to 90.27\%. It prunes 38.05M parameters and decreases MACs to 7.36G, modestly enhancing speedup by 2.40x. Random pruning shows a significant loss in performance and has a pruned test accuracy of 82.98\% and an accuracy drop of -13.58\%. L1-Norm and L2-Norm perform worse than the others when we consider accuracies of 80.85\% and 86.12\% and speed up metrics, respectively.

For 75\% pruning the DeiT model displays a huge reduction in the parameters and MACs reductions compared to 50\% on the original implementation. At 75\% pruning, the Hessian technique retains its advantage by showing highest performance of 75.89\% test score and -20.67\% accuracy loss. 
The behaviour of the DeiT model at an extreme pruning ratios changes rapidly. At 95\% pruning, performance degrades across all approaches. Hessian maintains its lead but only scores 54.41\% test accuracy, a decline of -42.15\%. Table~\ref{chapter7_tab4} also illustrates that under extreme pruning rates, the DeiT model has removed almost all the parameters and MACs are less than 1G for all models which is why their speed becomes even higher by decreasing  accuracy with larger margins. It is also clear that under extreme pruning conditions random selection performs even better than sophisticated methods. 

In summary, the Hessian approach regularly outperforms other pruning strategies, especially at low and moderate pruning levels (50\% and 75\%). This is because Hessian-based pruning uses first/second-order derivatives, which provide more detailed information on how each parameter affects the model's performance. This new information allows it to more effectively target and prune less significant parameters while conserving those critical to the model's structure and function. On the other hand, methods such as Taylor and random pruning, produce less consistent outcomes. Taylor pruning is based on gradient information and it struggles at higher pruning levels (75\% and 95\%) because it fails to capture complicated relationships of the network. Random pruning is less successful at lower pruning levels however it has surprising resilience at high pruning (95\%) since the randomisation helps avoid overfitting. Meanwhile, L1-Norm and L2-Norm pruning perform badly at all levels because they rely on magnitude which does not consider the complexities in the parameter of the DeiT model.

\begin{figure*}[!htb] 
\centering
\includegraphics[width=0.9\textwidth]{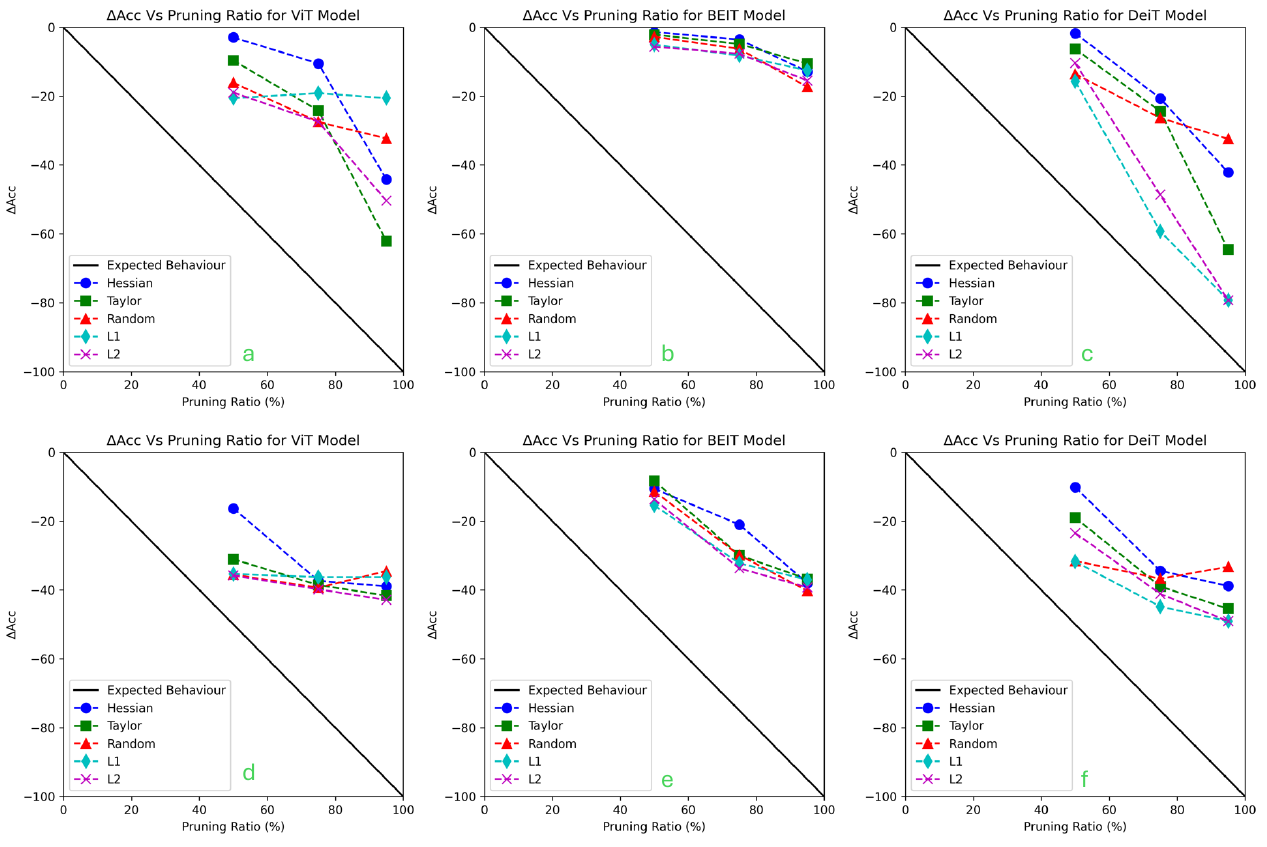} 
\caption{Relationship between pruning ratios and $\Delta$Acc, which represents the reduction rate in accuracy after pruning the models for PACS and Office-Home DG benchmarks: (a) ViT, (b) BeiT, and (c) DeiT, show analysis for PACS and (d) ViT, (e) BeiT, and (f) DeiT are for Office-Home}
\label{chapter7_fig6}
\end{figure*}

As another prospective of these results, we explore the relationship between the pruning ratios of the models and their resulting accuracy reduction. The graphs in Figure~\ref{chapter7_fig6} attempt to provide an explanation for the behaviour of each model (ViT, BeiT, DeiT) with respective pruning criteria and ratios. The x-axes in the graphs represent the pruning ratio (50, 75 or 95)\% and $\Delta$Acc presents the accuracy drop after pruning the models compared to baseline accuracy of same model which was shown earlier in Table~\ref{chapter7_tab1}. All the graphs show that overall with the increase in pruning rates the $\Delta$Acc also increases. However when we drop 50\% of any model using pruning methods its parameters and MACs also drop accordingly which is shown in Tables~\ref{chapter7_tab2}, \ref{chapter7_tab3} and \ref{chapter7_tab4}. Theoretically, by dropping 50\% of the connections, filters, links, or layers, accuracy should also be decreased by 50\%. Similarly, by increasing pruning to 75\% and 95\%, the expected accuracy drop should be close to those same ratios. In reality the $\Delta$Acc is far less than the expected behaviour. For example, in Figure~\ref{chapter7_fig6}(a), for Hessian the curve the $\Delta$Acc is at 2.94\% drop with 50\% removal of network and 44.14\% drop for the extreme case of a 95\% pruning ratio. This behaviour can be observed for other curves in Figure~\ref{chapter7_fig6}(a). The graphs in Figures~\ref{chapter7_fig6}(b) and \ref{chapter7_fig6}(c) also highlight the same trend of less accuracy dropping from the theoretical expectation which is a linear line in black colour in all the graphs.  

Figure~\ref{chapter7_fig6} gives us a general idea of performance of models under the influence of pruning. The second graph in Figure~\ref{chapter7_fig6}(b) has information about the BeiT model and illustrates that even in extreme conditions the $\Delta$Acc is less than compared to others models for all the curves and this also means that BeiT has better DG. Moreover,  Figure~\ref{chapter7_fig6}(a) displays that L1 norm has stable performance for all pruning ratios.


In summary, all the graphs show that the performances of models for each pruning criterion and following grouped structured pruning give better than expected behaviour and this behaviour may be caused by an interesting phenomena called ``grokking". Grokking is the phenomenon in which LLMs or any models are extensively trained, often to the point of overfitting but then their performance abruptly shifts into a stage of generalisation in which they begin to perform well on new and previously unexplored datasets. This transformation occurs unexpectedly and we still do not completely grasp the underlying mechanisms that drive it. It
is related to the concept of emergence which occurs when a complex system (such as a neural network or a large population) shows behaviours or features that its individual elements do not exhibit on their own\cite{smeaton2024understanding}.  Grokking is the computational equivalent of that.

\section{Conclusions}
This paper investigated domain generalisation using grouped structural pruning of vision transformers, focusing on ViT, BeiT, and DeiT models with the PACS and Office-Home benchmarks. The framework presented here is adaptable to various vision transformers and data distributions and introduced a hardware-friendly pruning approach based on importance scores. While the DeiT model showed superior performance with slightly larger trainable parameters and MACs, the BEIT model demonstrated consistent and robust generalisation across different pruning methods and ratios, validating its suitability for domain generalisation tasks. 

The study also revealed intriguing behaviours under varying pruning levels. Hessian-based selection excelled at moderate pruning levels (50\% and 75\%) due to its precision, while Taylor and random selection proved effective at high pruning rates (95\%), suggesting the potential of randomness to uncover high-performing sub-networks. Additionally, the phenomenon of grokking, observed in smaller models under prolonged training, indicated the role of over-training in achieving emergent generalisation behaviours. The BeiT model with Hessian pruning at 50\% offered the best trade-off, maintaining minimal performance loss (-1.42\%) while significantly enhancing speed (1.81×). These findings provide valuable insights for extending the framework to more complex benchmarks and exploring dynamic training strategies.

\section*{Acknowledgments}
This work was funded by Science Foundation Ireland through the Science Foundation Ireland Centre for Research Training in Machine Learning (ML-Labs) (SFI/)18/CRT/6183) and the Insight the SFI Research Centre for Data Analytics (SFI/12/RC/2289\_P2).

\bibliographystyle{IEEEtran}
\bibliography{references-fix}
\end{document}